\documentclass[letterpaper, 10 pt, conference]{ieeeconf}  
\usepackage[dvipsnames]{xcolor}
\usepackage{soul}

\IEEEoverridecommandlockouts                              

\usepackage{balance}
\usepackage{cite}
\usepackage{graphicx} 
\usepackage{flushend}
\begin{document}

\title{\LARGE \bf
The Effect of Trust and its Antecedents on Robot Acceptance 
}

\author{Katrin Fischer$^{1}$, Donggyu Kim$^{1}$, and Joo-Wha Hong$^{2}$
\thanks {$^{1}$ Katrin Fischer {\tt\small katrinfi@usc.edu} and Donggyu Kim {\tt\small donggyuk@usc.edu} are with the Annenberg School for Communication at the University of Southern California, Los Angeles.
$^{2}$ Joo-Wha Hong {\tt\small joowhaho@marshall.usc.edu} is with the Marshall School of Business at the University of Southern California, Los Angeles.
  }
}
\maketitle
\begin{abstract}
As social and socially assistive robots are becoming more prevalent in our society, it is beneficial to understand how people form first impressions of them and eventually come to trust and accept them. This paper describes an Amazon Mechanical Turk study ($n = 239$) that investigated trust and its antecedents trustworthiness and first impressions. Participants evaluated the social robot Pepper’s warmth and competence as well as trustworthiness characteristics ability, benevolence and integrity followed by their trust in and intention to use the robot. Mediation analyses assessed to what degree participants’ first impressions affected their willingness to trust and use it. Known constructs from user acceptance and trust research were introduced to explain the pathways in which one perception predicted the next. Results showed that trustworthiness and trust, in serial, mediated the relationship between first impressions and behavioral intention.
\end{abstract}
\section{Introduction \& Related Work}

Trust plays an important role in human relationships and is similarly important in establishing relationships between humans and robots. It is “the confidence that one will find what is desired from another, rather than what is feared" \cite{Deutsch1973} and is crucial in the presence of (perceived) risk, e.g. when another’s abilities or actions cannot be foreseen \cite{Riegelsberger2003}. Consistent and predictable behaviors as well as trust-building characteristics such as dependability are cornerstones that make up trust in human relationships \cite{Rempel1985}. Research on human-robot trust has generated a sizeable body of literature in which multiple definitions and frameworks of trust exist. For instance, Lee \& See applied Rempel et al.'s \cite{Rempel1985} dimensions of human trust to interactions with robots in terms of their performance, behavior determining operations (algorithms) and the degree to which they are used within their designer’s intent \cite{Lee2004}. Other research has considered the multidimensional nature of trust with a focus on gains and losses in human-robot trust due to robot behaviour change \cite{10.5555/3378680.3378816}. We follow the model of Mayer et al. \cite{mayer1995integrative} who argued that trust is built through evaluating trustworthiness characteristics of the trustee, which has been shown to apply to HRI contexts by evaluating the robot's ability, integrity and benevolence \cite{Kim2020}.

Trustworthiness is also connected to social dimensions of human perceptions of robots such as warmth and competence which vary based on robot appearance \cite{Carpinella2017}. According to the stereotype content model \cite{Fiske2007}, these first impressions help us determine whether a new acquaintance is likely to constitute a friend or a foe. Warmth judgements (trustworthiness, helpfulness, perceived intent) occur first and decide affective and behavioral reactions, whereas perceived competence ascertains to what extent the other can act on their motives (perceived ability, efficiency, intelligence). Competence and warmth stereotypes predict emotions, which directly predict behaviors \cite{Cuddy2007} and they apply to people as well as to social robots \cite{ByronReeves2020}. There is a relationship between warmth and trust in that somebody that is perceived to be warm is simultaneously considered trustworthy, or its inverse, cold and untrustworthy \cite{Fiske2007}. 

Trust plays a crucial role in HRI \cite{Nam2020,Hancock2011} and has been recognized as a factor that not only predicts the quality of the interaction, but also how willing people are to use social robots for certain tasks \cite{Naneva2020,salem2015evaluating,Salem2015}. Among the most commonly employed models to assess use and acceptance of new technologies are TAM (Technology Acceptance Model) \cite{Davis1989} and UTAUT (Unified Theory of Acceptance and Use of Technology) \cite{Venkatesh2003}. TAM studies can explain approximately 50\% of the variance in technology acceptance outcomes while UTAUT has been found to produce an adjusted $R^2$ of over 69\% \cite{Venkatesh2003}. Its outcome, intention to use, is relevant to HRI research as a predictor of robot acceptance, on which, together with trust, the success of socially assistive robots depends \cite{Mataric2016}. Past research has looked at trust and acceptance in HRI, but not explored the specific pathways of their relationship.

\textbf{RQ1}. How do the above identified constructs of trust, trustworthiness, warmth, competence, and intention to use social robots interrelate? 

\textbf{RQ2}. What are the antecedents of trust? Of social robot acceptance?

\section{Method}

An online survey was disseminated through Amazon Mechanical Turk. A total of 239 participants (86 females, 153 males) passed the attention check and completed all questionnaires. Empirical estimates of sample sizes needed for 0.8 power in mediation analysis confirm that this sample size is adequate to detect small to medium effects when using Hayes’ PROCESS version 3+ utilizing percentile bootstrap confidence intervals as the default \cite{Fritz2007}.

After consenting to participate in the study, participants were exposed to an image and textual description of the social robot Pepper and filled out demographic questions regarding their age ($M = 35.15$), gender (86 female, 153 male), education ($M = 4.75$ years of college/university) and familiarity with robots ($M = 3.56$ on a 7-point Likert scale) as well as warmth and competence questions, UTAUT assessments and trust and trustworthiness questionnaires. All responses were collected on 7-point Likert scales.

\begin{figure}[ht] \centering \includegraphics[width=0.3\textwidth]{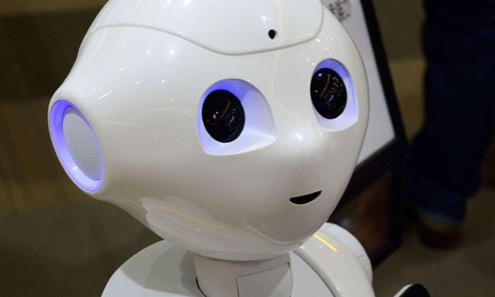} \caption{The social robot Pepper.}
\label{pilot}
\end{figure}

The stereotype content model \cite{Fiske2007} was used to measure first impressions as it has been successfully applied to human perceptions of social robots \cite{ByronReeves2020}. The warmth and competence items utilized the statement “As viewed by society, how ... are social robots?”. The warmth dimension contained four items: tolerant, warm, good natured, and sincere ($\alpha = 0.74$). Competence was comprised of five items: Competent, confident, independent, competitive, and intelligent ($\alpha = 0.78$). 
The UTAUT construct behavioral intention \cite{Venkatesh2003} consisted of three items (example: “I intend to use the robot”, $\alpha = 0.71$).
We utilized a measure of trust that considered its relationship with trustworthiness and provided assessments of robot ability, benevolence and integrity \cite{Kim2020}. Trust was thereby measured directly via 10 questions ($\alpha = 0.84$), of which one example item was “I would rely on the robot without hesitation”. Trustworthiness was measured via the dimensions of ability (six items, example: “I feel very confident about the robot’s skills”, $\alpha = 0.80$),  benevolence (five items, example: “the needs and desires of others are very important to the robot”, $\alpha = 0.82$) and integrity (five items, example: “sound principles seem to guide the robot’s behavior”, $\alpha = 0.80$). An aggregate of these three subscales produced the trustworthiness index ($\alpha = 0.85$).
\section{Analysis \& Results}

The research questions were investigated via a series of mediation analyses. Mediation analysis informs the relation between two variables by explaining how they are related \cite{Fairchild2009}. The analysis models how a mediator variable transmits the effect of an independent variable X on a dependent variable Y. Path analysis is an extension of multiple regression that can examine chains of influence as well as several dependent variables \cite{Streiner2005}. The PROCESS macro for R (version 4.0.2) was used for all mediation analyses \cite{Hayes2022}. Mediation models tested the relationship between first impressions, trustworthiness, trust and intention to use. First impressions were split into its components warmth and competence and as these are thought to occur first, entered  in turn as independent variable X for each model. Mediators were explored to predict the outcomes trust and intention use. A final model integrated all four concepts of interest in a serial parallel mediation.

\subsection{Simple Mediation}

The effect of warmth on trust was fully mediated via trustworthiness (see Fig. 2). The regression coefficient between warmth and trustworthiness as well as the regression coefficient between trustworthiness and trust were significant. The indirect effect was (.55) * (.73) = .40. The significance of the indirect effect was tested using 5,000 bootstrap samples and the 95\% confidence interval ranged from 0.311 to 0.500. This means that 2.5\% of the 5,000 bootstrap estimates were smaller than 0.311 and 2.5\% were larger than 0.500. As this confidence interval does not include zero, we can reject the null hypothesis that $_{T}a_{T}b = 0$ and conclude that the indirect effect is statistically significant. There was a significant total effect ($c = 0.48$, 95\% $CI$ 0.385 to 0.581), but no significant direct effect ($c’ = 0.08$, 95\% $CI$ -0.005 to 0.165).

\begin{figure}[ht] \centering \includegraphics[width=0.5\textwidth]{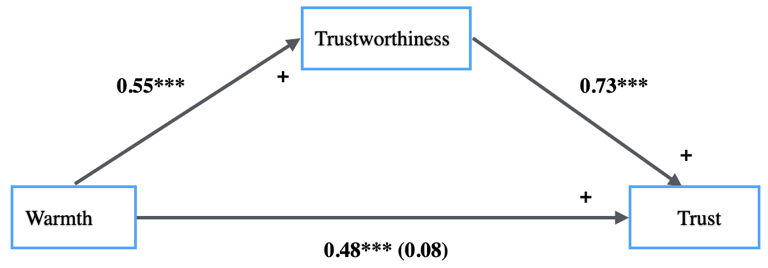} \caption{Simple Mediation Results (Warmth - Trustworthiness - Trust).}
\label{pilot}
\end{figure}

Similarly, results showed that competence has a significant total effect on trust ($c = 0.47$, $p < 0.001$). Analyzing the indirect effects reveals that trustworthiness significantly mediates the relationship between competence and trust, $ab = 0.45$, $p < 0.01$ (95\% $CI$ 0.352 to 0.554). Competence positively affects trustworthiness ($a = 0.59$, $p < 0.001$) and trustworthiness, in turn, positively affects trust ($b = 0.77, p < 0.001$). After accounting for the mediating role of trustworthiness, the direct effect is not significant, indicating that the total effect is fully explained by the mediator (complete mediation). These findings provide some evidence that people who perceive a social robot as competent are more likely to trust it as they tend to also perceive its characteristics as trustworthy. However, people’s ratings of competence do not contribute to trust beyond what is accounted for by trustworthiness.

\subsection{Parallel Mediation}

Two parallel mediator models were designed to test the impact of the mediator trustworthiness in more detail. As trustworthiness is an index made up of the three subscales ability, benevolence and integrity, these constructs were next entered as parallel mediators. The indirect effect of warmth on trust through ability (X → M1 → Y), estimated as $a_1 * b_1 = 0.16$ was significant (95\% $CI$ 0.089 to 0.241), as was the indirect effect of warmth on trust through benevolence (X → M2 → Y) estimated as $a_2 * b_2 = 0.13$ (95\% $CI$ 0.049 to 0.239). However, integrity (X → M3 → Y) did not significantly mediate the relationship between warmth and trust ($a_3b_3 = 0.10$, 95\% $CI$ -0.002 to 0.218). Both the total effect ($c = 0.48$) and the direct effect ($c’ = 0.09$) were statistically significant. 
A similar parallel mediator model, with the first impression of robot competence as independent variable instead of its warmth, confirmed this trend (see Fig. 3). 

\begin{figure}[ht] \centering \includegraphics[width=0.4\textwidth]{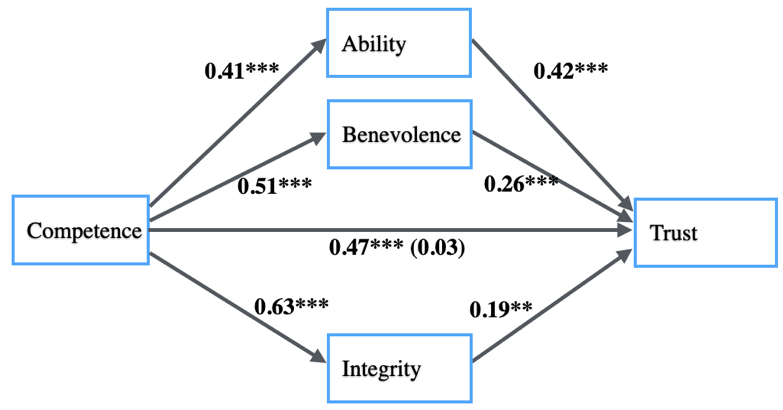} \caption{Parallel Mediation Results (Competence - Trustworthiness constructs - Trust).}
\label{pilot}
\end{figure}

The indirect effect of competence on trust through ability (X → M1 → Y), estimated as $a_1 * b_1 = 0.18$ was significant (95\% $CI$ 0.104 to 0.254), as was the indirect effect of competence on trust through benevolence (X → M2 → Y) estimated as $a_2 * b_2 = 0.14$ (95\% $CI$ 0 0.053 to 0.248). However, even though the estimates for the individual legs $a_3$ and $b_3$ were significant (see Fig. 3), integrity (X → M3 → Y) did not significantly mediate the relationship between competence and trust as the confidence interval included zero ($a_3b_3 = 0.12$, 95\% $CI$ -0.007 to 0.248). The total effect ($c = 0.47$) was significant, however, the direct effect ($c’ = 0.03$) was not. As integrity did not mediate the relationship between first impressions and trust when entered in parallel with ability and benevolence, a serial model will explore whether integrity in serial with ability and benevolence will provide a significant indirect effect as there is literature indicating this possibility \cite{Kim2020}.

\subsection{Serial Parallel Mediation}

Extending the previous two parallel mediator models, a serial model with integrity predicting ability and benevolence predicting trust was run. The direct effect of X on Y was fixed to zero. The indirect effect of warmth on trust through integrity and ability (X → M1 → M2 → Y), estimated as $a_1 * d_{21} * b_2 = 0.59 * 0.54 * 0.50 = 0.16$ was significant (95\% $CI$ 0.095 to 0.230). The indirect effect of warmth on trust through integrity and benevolence (X → M1 → M3 → Y), estimated as $a_1 * d_{31} * b_3 = 0.59 * 0.80 * 0.39 = 0.18$ was significant (see Fig. 4). The contrast was not significant. 

\begin{figure}[ht] \centering \includegraphics[width=0.5\textwidth]{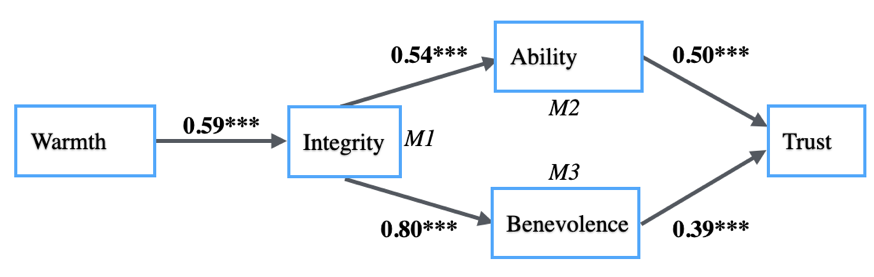} \caption{Serial Mediator Model (Warmth - Integrity - Ability/Benevolence - Trust).}
\label{pilot}
\end{figure}

A similar model with competence as the independent variable showed comparable results. The indirect effect of competence on trust through integrity and ability ($a_1 * d_{21} * b_2 = 0.17$) as well as the indirect effect through integrity and benevolence ($a_1 * d_{31} * b_3 = 0.19$) were significant.

Finally, in order to express the relationship between first impressions, trustworthiness, trust and behavioral intention in one comprehensive model, a serial parallel mediation model was devised using PROCESS model 80 via B matrix specification. The direct effect of X on Y was fixed to zero. The indirect effect of competence on behavioral intention through ability and trust (X → M1 → M4 → Y), estimated as $a_1 * d_{41} * b_4 * = 0.41 * 0.43 * 0.61 = 0.11$ was significant (95\% $CI$ 0.060 to 0.171). 

\begin{figure}[ht] \centering \includegraphics[width=0.5\textwidth]{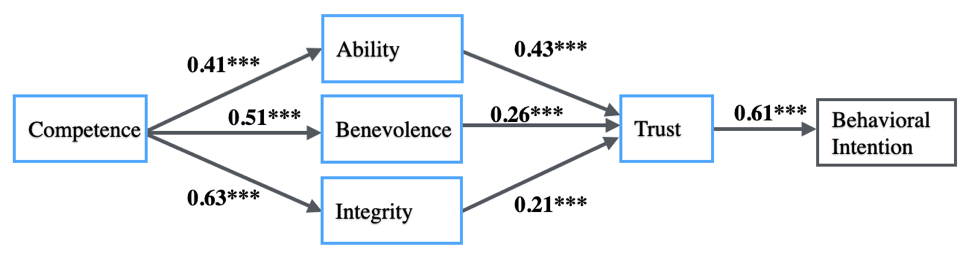} \caption{Serial Parallel Mediator Model (Competence - Trustworthiness constructs - Trust - Behavioral intention).}
\label{pilot}
\end{figure}

Similarly, the indirect effects of competence on behavioral intention through benevolence and trust (X → M2 → M4 → Y), estimated as $a_2 * d_{42} * b_4 = 0.08$ and competence on behavioral intention through integrity and trust (X → M3 → M4 → Y), estimated as $a * d_{43} * b_4 * = 0.08$ were significant (see Fig. 5). A serial model with the first impression warmth as independent variable provided similar results. All indirect effects were significant ($a_1 * d_{41} * b_4 = 0.10$, $a_2 * d_{42} * b_4 = 0.08$, $a_3 * d_{43} * b_4 = 0.08$). 

\section{Discussion \& Conclusion}

This study ($n=239$) was conducted to explore the impact of first impressions and trustworthiness characteristics on social robot trust and acceptance. The results of the mediation analysis show that while there was no direct effect of first impressions on trust, the indirect effects of both warmth and competence on trust via the mediator trustworthiness were significant. Separating trustworthiness into its sub scales ability, benevolence and integrity run as parallel mediation confirmed that both ability and benevolence, but not integrity, produced significant indirect effects on trust. Other research \cite{Kim2020} has documented this phenomenon and suggested that of these three sub scales, integrity only predicts trust via ability and benevolence. A serial parallel model placing integrity before ability and benevolence, which in turn predicted trust, confirmed this. Finally, a serial parallel mediation model including first impressions, trustworthiness constructs, trust and behavioral intention per the UTAUT model was devised and analyzed. All indirect effects were significant, indicating that participants whose first impression of the social robot was positive were likely to intent to use if they also perceived it as trustworthy and had trust in it. Notably, in this final model integrity was placed in parallel with ability and benevolence and produced a significant effect predicting intention to use via trust, while a prior model showed that it predicted a trust outcome only in serial with ability and benevolence, indicating that predicting intention to use may rely on all three subscales to a higher degree than trust does. This model gives evidence that stereotypical first impressions, trustworthiness, trust and use behavior are closely related and informed by each other. We recommend further investigations into how real-world interactions may influence the reported results and how contextual changes and different user segments may affect the role of trustworthiness characteristics.

\bibliography{bibliography}

\balance
\end{document}